\documentclass{article}

\usepackage[multiple]{footmisc}

\usepackage[nonatbib]{neurips_2022}
\nolinenumbers
\usepackage[utf8]{inputenc} 
\usepackage[T1]{fontenc}    
\usepackage{url}            
\usepackage{booktabs}       
\usepackage{amsfonts}       
\usepackage{nicefrac}       
\usepackage{microtype}      
\usepackage{xcolor}         
\usepackage[pagebackref,breaklinks,colorlinks]{hyperref}
\usepackage{url}       
\usepackage{booktabs}     
\usepackage{amsfonts}   
\usepackage{amsmath}
\usepackage{nicefrac}      
\usepackage{microtype}
\usepackage{multirow}
\usepackage{wrapfig}
\usepackage{threeparttable}
\usepackage{graphicx}
\usepackage{xcolor}         
\usepackage{enumitem}

\newcommand{\sref}[1]{Section.~\ref{#1}}

\newcommand{\figureref}[1]{Figure~\ref{#1}}

\newcommand{\tableref}[1]{Table~\ref{#1}}

\let\oldFootnote\footnote
\newcommand\nextToken\relax

\renewcommand\footnote[1]{%
    \oldFootnote{#1}\futurelet\nextToken\isFootnote}

\newcommand\isFootnote{%
    \ifx\footnote\nextToken\textsuperscript{,}\fi}

\title{1st Place Solution to NeurIPS 2022 Challenge on Visual Domain Adaptation}

\author{%
   Daehan Kim\thanks{This author was supported by SI Analytics.}\\
   Hanbat National University\\
   \texttt{daehan.kim@edu.hanbat.ac.kr}
   \And
   Minseok Seo\\
   SI Analytics\\
   \texttt{minseok.seo@si-analytics.ai}\\
   \And
   YoungJin Jeon\\
   SI Analytics\\
   \texttt{youngjin.jeon@si-analytics.ai}\\
   \And
   Dong-Geol Choi\thanks{Corresponding author}\\
   Hanbat National University\\
   \texttt{dgchoi@hanbat.ac.kr}\\
}
\begin{document}

\maketitle

\begin{abstract}
The Visual Domain Adaptation(VisDA) 2022 Challenge calls for an unsupervised domain adaptive model in semantic segmentation tasks for industrial waste sorting.
In this paper, we introduce the \texttt{SIA\_Adapt} method, which incorporates several methods for domain adaptive models.
The core of our method in the transferable representation from large-scale pre-training.
In this process, we choose a network architecture that differs from the state-of-the-art for domain adaptation.
%
%
After that, self-training using pseudo-labels helps to make the initial adaptation model more adaptable to the target domain.
Finally, the model soup scheme helped to improve the generalization performance in the target domain.
Our method \texttt{SIA\_Adapt} achieves 1st place in the VisDA2022 challenge.
The code is available on \url{https://github.com/DaehanKim-Korea/VisDA2022_Winner_Solution}.
\end{abstract}

\section{Introduction} \label{section:introduction}
Deep learning-based models perform well on training data distributions(source) due to bias, but otherwise(target) perform poorly~\cite{hoffman2013one, wang2018deep}.
This phenomenon is common in visual recognition tasks such as classification, object detection, and semantic segmentation.
Until recently, research on domain generalization(DG) and domain adaptation(DA) related to domain transfer has been proposed as a method to alleviate this phenomenon.
The VisDA 2022 challenge is closely related to DA that can access unlabeled target domains.
Categorically, it is unsupervised domain adaptation(UDA) in semantic segmentation.

A recent study~\cite{kim2022unified} shows that pre-trianing has a significant impact on downstream tasks such as domain adaptation. 
Also, these impacts include network architecture design, size, pre-training loss, and datasets.
Therefore, we known that the pre-training step is important in designing a domain adaptive model, and the domain adaptation step to the target domain is still helpful.

Inspired by this study, we propose \texttt{SIA\_Adapt} as a domain adaptive model for industrial waste sorting.
\texttt{SIA\_Adapt} uses DAFormer~\cite{hoyer2022daformer} framework as the baseline, but not the Mix Transformers(MiT) backbone of SegFormer~\cite{xie2021segformer}.
SegFormer is designed for semantic segmentation tasks and achieves compelling performance on various datasets such as ADE20K~\cite{zhou2017scene}, Cityscapes~\cite{cordts2016cityscapes}, and COCO-Stuff~\cite{caesar2018coco}.
However, we choose ConvNeXt~\cite{liu2022convnet} backbone for stronger domain adaptation and transfer of large-scale prior knowledge from ImageNet-22K~\cite{russakovsky2015imagenet}.
It is noteworthy that even if the domain adaptation step is omitted, simply changing the network architecture and using a large-scale prior knowledge can achieve remarkable performance without access to the target domain samples.

After that, we apply self-training using pseudo-labels and model soup~\cite{wortsman2022model} schemes to maximize performance in the target domain.
%

%
As a result, \texttt{SIA\_Adapt} achieves 1st place in both mIoU and Acc in the VisDA2022 challenge.

\section{Method} \label{section:method}
In this section, each component of \texttt{SIA\_Adapt} is described.
\sref{section:baseline} describes the baseline framework of \texttt{SIA\_Adapt}.
\sref{section:inital_adaptive_model} describes core methods applied to improve the performance of the initial adaptive model in the target domain.
\sref{section:self_training} describes a pseudo-labeling process that uses an initial adaptive model to generate incomplete labels for unlabeled target datasets.
Then, \sref{section:model_soups} describes a model soup method to improve domain generalization performance using multiple-finetuned models generated through self-training using pseudo-label.

\begin{figure*}[!t]

  \centering
  \includegraphics[width=1.0\linewidth]{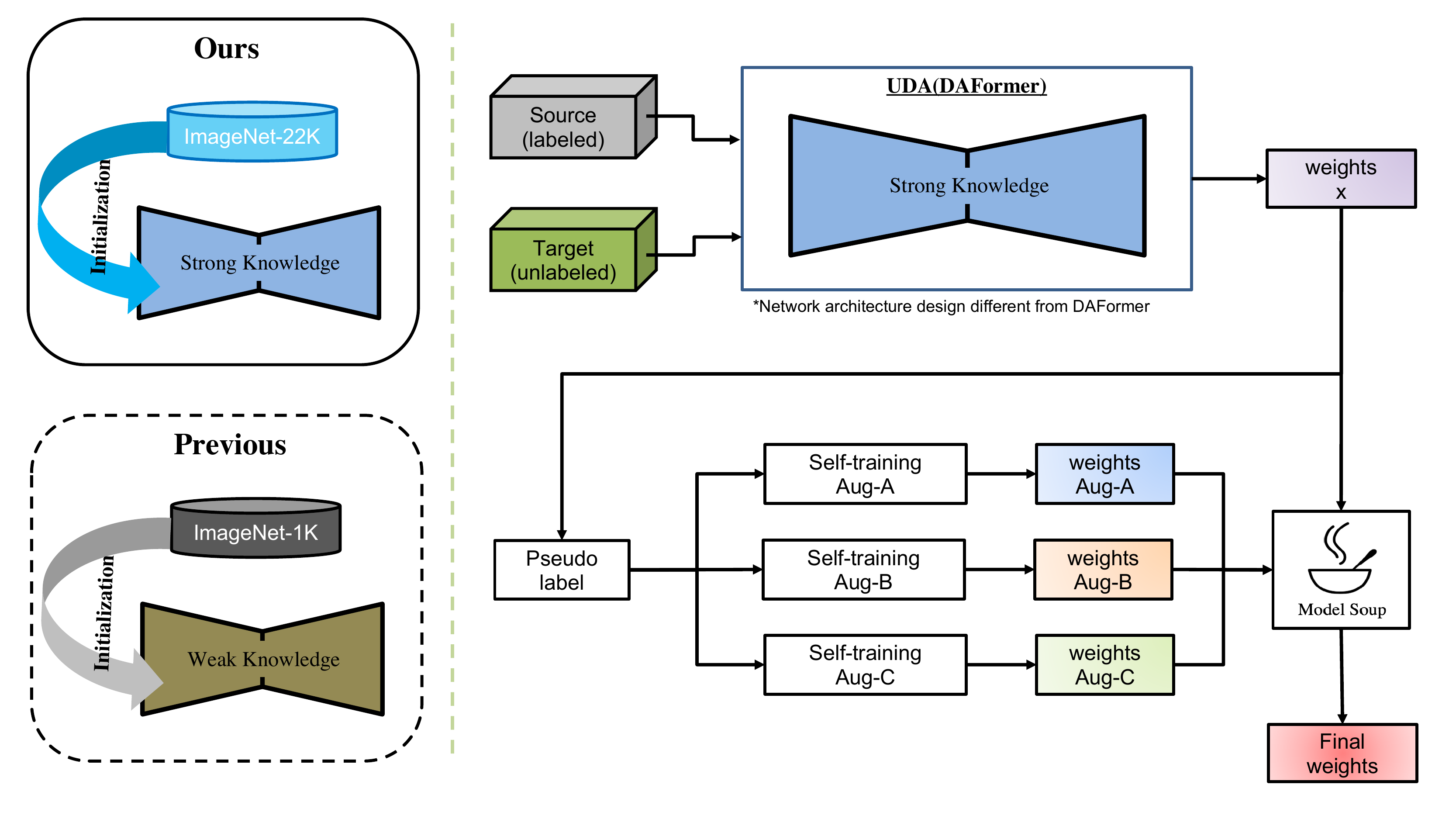}
  \caption{
  Ovierview of \texttt{SIA\_Adapt}.
  }
  \label{figure:figure1}
\end{figure*}

\subsection{Overview} \label{section:overview}
\figureref{figure:figure1} is an overview of \texttt{SIA\_Adapt}.
In the first step of \texttt{SIA\_Adapt}, unsupervised domain adaptation training is performed on a labeled source dataset $\mathcal{D}_{s}$ and an unlabeled target dataset $\mathcal{D}_{t}$.
After UDA training, we get a initial adaptive model $\mathcal{G}_{init}$.
Before the next step, pseudo-labels are generated for the target domain with the corresponding model.
Based on the pseudo-label, self-training is performed using three different augmentation.
The multiple-finetuned model $\mathcal{G}_{aug\_a}$, $\mathcal{G}_{aug\_b}$, $\mathcal{G}_{aug\_c}$ obtained through self-training is used as a model soup recipe.
Finally, it is weighted averaged through the greedy soup to produce the final adaptive model $\mathcal{G}_{final}$.

\subsection{Baseline framework of \texttt{SIA\_Adapt}}
\label{section:baseline}

Unsupervised domain adaptation for semantic segmentation~\cite{tsai2018learning, hoffman2018cycada, vu2019advent, yi2022pt4al, wang2021domain} has been studied for a considerable period based on ResNet~\cite{he2016deep} backbone DeepLabv2~\cite{chen2017deeplab}.
However, DAFormer points out the absurdity of using the outdated network architecture and proposes a new architecture for domain adaptation.
DAFormer network architecture consists of a transformer-based encoder and a multi-level context-aware fusion decoder.
In addition, a simple but critical training strategy for domain adaptation is applied.

\begin{figure*}[!t]
  \centering
  \includegraphics[width=1.0\linewidth]{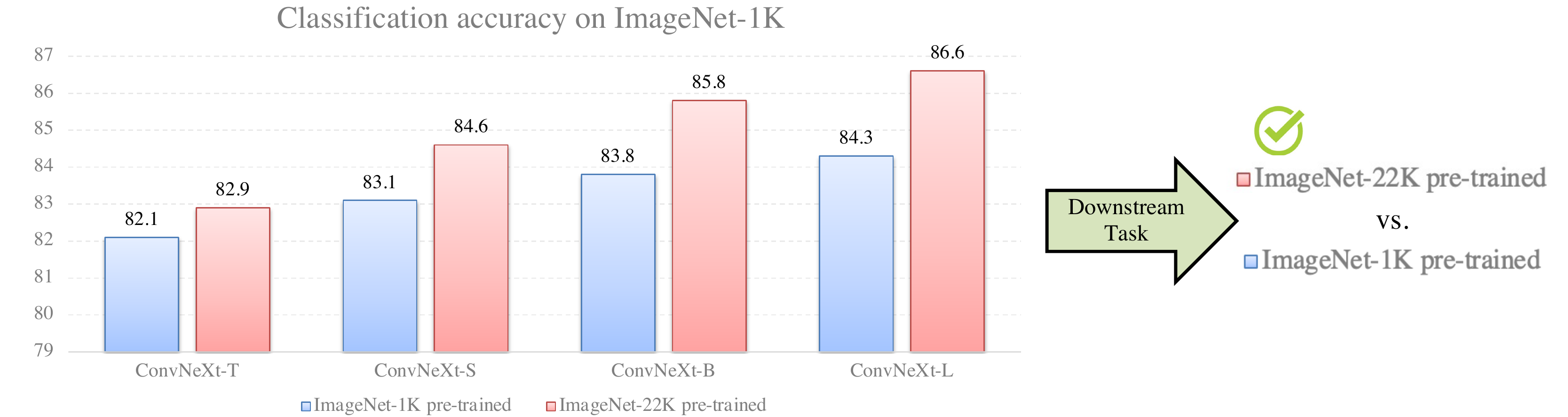}
  \caption{
  ImageNet-1K vs. ImageNet-22K accuracy comparison of ConvNeXt architecture. Even for downstream tasks such as domain adaptation, ImageNet-22K pre-trained weights potentially improves transfer.\vspace{-2.5mm}
  }
  \label{figure:figure2}
\end{figure*}

The factors that affect the performance improvement of DAFormer are as follows: 1. Mix Transformers architecture designed for semantic segmentation, 2. Rare Class Sampling, 3. Thing-Class ImageNet Feature Distance, 4. Learning Rate Warmup for UDA.

While following most training strategy of DAFormer, we have empirically found that Rare Class Sampling leads to performance drop in the VisDA2022 challenge scenario, and we exclude it from training.
We expect the co-occurrence relationship between minor and major classes to be high in urban traffic scenarios(e.g. bicycles and people) such as cityscapes, gta5 and synthia, while low in VisDA2022 scenarios(e.g. rigid\_plastic and cardboard).

\subsection{Initial adaptive model of \texttt{SIA\_Adapt}}
\label{section:inital_adaptive_model}

We use a state-of-the-art framework of unsupervised domain adaptation for semantic segmentation as a baseline.
However, according to \textit{Kim, Donghyun, et al.}~\cite{kim2022unified}, pre-training, network architecture design, size, etc. influence the domain adaptation task.
As shown in ~\figureref{figure:figure2}, it is a wise choice to use ImageNet-22k pre-trained weights because the performance trends in ImageNet-1K are also maintained for downstream tasks.
In other words, there are still plenty of improvements left in DAFormer.

Therefore, we use the initial weights trained with ImageNet-22k for initial adaptive training.
%
However, so far there are no publicly available ImageNet-22k pre-trained weights on the MiT backbone, so we change the backbone architecture design to ConvNeXt.
ConvNeXt backbones are competitive compared to other modern backbone families~\cite{arnab2021vivit, touvron2021training, liu2021swin}.
Also, there are publicly available Imagenet-22k pre-trained weights.
Furthermore, a deeper network can lead to better domain adatation performance, so we use a moderately deep ConvNeXt-L for accuracy and efficiency.
In these network architecture modifications, the decoder design follows DAFormer.

\subsection{Self-training with pseudo-label} \label{section:self_training}
Self-training with pseudo-labels is a simple but effective approach widely used in tasks such as unsupervised domain adaptation~\cite{zou2018unsupervised, mei2020instance, tranheden2021dacs, zhang2021prototypical, seo2022bag} and semi-supervised domain adaptation~\cite{chen2021semi}.
To apply the self-training approach, pseudo-labels for unlabeled target domain were generated with initial adaptive model in \sref{section:inital_adaptive_model}.

Although the initial adaptive model shows remarkable performance for the target domain samples, noise labels may be generated.
Therefore, we exclude pixel with confidence less than 0.9 from self-training because they are closely label-noise and drop the generalization performance of the model.
We empirically confirm that setting the confidence threshold to 0.9 contributes to performance improvement and use it.
Self-training is trained on an initial adaptive model, various augmentations are used individually, and the multiple-finetuned model is then used as a model soup recipe.

\subsection{Model soups} \label{section:model_soups}
Recently proposed, Model Soups achieves state-of-the-art in difficult benchmarks such as ImageNet-A~\cite{hendrycks2021natural}, ImageNet-R~\cite{hendrycks2021many}, and Imagenet-Sketch~\cite{wang2019learning}.
Also, Model Soup has been included as a winning solution in prestigious challenges such as AI CITY CHALLENGE (CVPR2022)~\cite{yang2022box}, Benchmarking Multi-Target Tracking (CVPR2022)~\cite{seo2022bag}, and Seasons in Drift Challenge (ECCV2022)~\cite{johansen2022chalearn}, and has proven its effectiveness in various tasks.
The strength of Model Soup is that it can produce a better model by averaging the weights of multiple fine-tuned solutions without additional computation during the inference process.
Therefore, we adopt Model Soup schemes in our solution to prevent over-fitting to pseudo-labeled target datasets and to improve model performance.
%
Consequently, the initial adaptive model in \sref{section:inital_adaptive_model} and the three models generated in \sref{section:self_training} are used as model soup recipes and are weighted averaged.

\section{Experiments}
\subsection{Implementatioin Details} \label{section:implementation_details}
\paragraph{Datasets}
We use ZeroWastev1 as the labeled source domain and ZeroWastev2 as the unlabeled target domain according to the VisDA 2022 challenge rule\footnote{https://ai.bu.edu/visda-2022/}.
In training, SynthWaste and SynthWaste-aug is not used. 
Also, in the evaluation phase, it is performed on the ZeroWastev2 test set.

\paragraph{Training}
We trained the model using the DAFormer official code\footnote{https://github.com/lhoyer/DAFormer}\footnote{https://github.com/dbash/visda2022-org}.
IN-22K pre-trained weights for ConvNeXt-L are publicly available\footnote{https://github.com/facebookresearch/ConvNeXt}.
Additionally, NVIDIA RTX8000 GPU x1 is used and all hyperparameters except for rare class sampling strictly follow the original code.
Note that, we trained 40,000 iterations to the initial adaptive model and 10,000 iterations to the fine-tuned model.

\paragraph{Fine-tuninig}
We followed the model soup recipe to change the data augmentation type during self-training and Exponential Moving Average (EMA) is not used.
Augmentation used is \texttt{PhotoMetricDistortion} implemented by mmseg\footnote{https://github.com/open-mmlab/mmsegmentation}, \texttt{GaussNoise} and \texttt{RandomGridShuffle} implemented by albumentations\footnote{https://github.com/albumentations-team/albumentations}.

\begin{table*}[h]
\centering
\resizebox{0.88\textwidth}{!}{
 \def\arraystretch{1.1}

 \begin{tabular}{l|c|c|c|c|c|c}
    \hline \hline
 
    {Method} &  {Background} & Rigid Plastic & Cardboard & Metal & Soft Plastic & mIoU\\
    \hline
    {Weight(UDA)} &  {92.65} & \textbf{48.40} & 65.26 & 34.38 & 52.91 & 58.72\\
    \hline
    {Weight(PhotoMetricDistortion)} &  {92.75} & 47.69  & 65.57 & 33.96 & 54.60  & 58.91\\
    {Weight(GaussNoise)} &  {92.75} & 47.53 & 65.59 & 33.92 & 55.04 & 58.97 \\
    {Weight(RandomGridShuffle)} &  {92.73} & 47.47 & 65.43 & 34.21 & 54.70 & 58.91 \\
    \hline
    {Weight(Model Soup)} &  {\textbf{92.80}}& 48.14 & \textbf{65.80} & \textbf{35.27} & \textbf{55.11} & \textbf{59.42}\\


    \hline
    \hline
    \end{tabular}
    }
    \caption{Quantitative evaluation of our method on the ZeroWastev2 test set}
    \label{table:A.1}

\end{table*}

\begin{table*}[h]
\centering
\resizebox{0.88\textwidth}{!}{
 \def\arraystretch{1.1}

 \begin{tabular}{c|c|c|c|c|c}
    \hline \hline
 
    \multirow{2}{*}{Rank} &  \multirow{2}{*}{Team Name} & \multicolumn{2}{c}{UDA} & \multicolumn{2}{c}{Source Only}\\
    \cline{3-6}
    & & mIoU & Acc & mIoU & Acc\\
    \hline
    1st & \textbf{\texttt{SI Analytics}} & \textbf{59.42} & \textbf{93.18} & \textbf{56.46} & \textbf{93.38} \\
    2nd & \texttt{Pros} & 55.46 & 92.59 & 38.32 & 90.81\\
    3rd & \texttt{BIT-DA} & 54.38 & 91.80 & 47.22 & 92.14\\
    4th & \texttt{TianQing} & 51.74 & 92.11 & 50.54 & 92.30\\
    - & \texttt{Baseline(DAFormer)} & 52.26 & 91.20 & 45.40 & 91.64\\

    \hline
    \hline
    \end{tabular}
    }
    \caption{VisDA2022 challenge final leaderboard results.}
    \label{table:A.2}
\end{table*}

\subsection{Challenge Results} \label{section:challenge_results}
The VisDA 2022 Challenge uses mIoU and Acc of UDA task as performance metrics.
The detailed performance of our method is shown in ~\tableref{table:A.1}.
Also, as shown in ~\tableref{table:A.2}, our method was evaluated on the ZeroWastev2 test set and achieved 1st place with 59.42 mIoU and 93.18 Acc, a large gap from 2nd place.
Notably, our method achieves 56.46 mIoU and 93.38 Acc in Source Only without accessing the target domain, still maintaining 1st place.

\section{Conclusions}
In this challenge, we proposed \texttt{SIA\_Adapt} for unsupervised domain adaptation in semantic segmentation that achieves 59.42 mIoU on the ZeroWastev2 dataset and 1st place on the leaderboard.
Our proposed method does not require large memory and can perform all training processes in one day.
We have confirmed that design for network architecture, size, etc. and pre-trained weights are important for domain adaptation.
Therefore, we plan to consider this and explore more efficient domain adaptation methods.





{\small
\bibliographystyle{plain}
\bibliography{egbib}
}

\end{document}